\documentclass{article}
\PassOptionsToPackage{numbers, compress}{natbib}
\usepackage[preprint]{neurips_2025}

\usepackage[utf8]{inputenc} % allow utf-8 input
\usepackage[T1]{fontenc}    % use 8-bit T1 fonts
\usepackage{hyperref}       % hyperlinks
\usepackage{url}            % simple URL typesetting
\usepackage{booktabs}       % professional-quality tables
\usepackage{amsfonts}       % blackboard math symbols
\usepackage{nicefrac}       % compact symbols for 1/2, etc.
\usepackage{microtype}      % microtypography
\usepackage{xcolor}         % colors
\usepackage{graphicx}
\usepackage{amsmath} 
\usepackage{enumitem}

\title{DIAL: Decoupling Intent and Action via Latent World Modeling for End-to-End VLA}

\author{
 Yi Chen$^{1}$ \quad 
 Yuying Ge$^{2\dagger}$ \quad 
 Hui Zhou$^{2}$ \quad 
 Mingyu Ding$^{3}$ \quad 
 Yixiao Ge$^{2}$ \quad 
 Xihui Liu$^{1\dagger}$ \\
 $^1$The University of Hong Kong \quad
 $^2$XPENG Robotics \quad 
 $^3$University of North Carolina at Chapel Hill\\
 \small{\color{blue} \url{https://xpeng-robotics.github.io/dial}}
}

\begin{document}

\maketitle

\let\thefootnote\relax\footnotetext{$\dagger$ Corresponding authors.}

\begin{figure}[!htbp]
    \centering
    \includegraphics[width=1.0\textwidth]{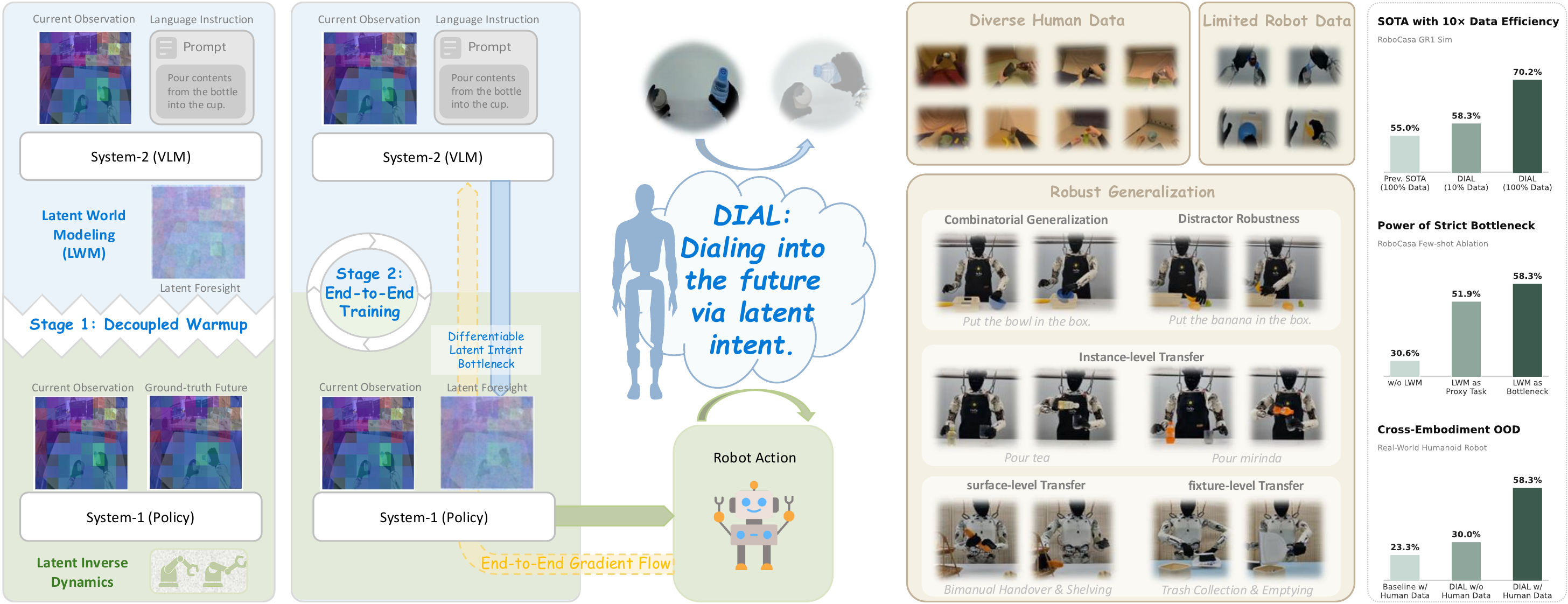}
    \caption{
    % \textbf{Overview of the DIAL Framework.} 
    DIAL bridges  high-level decision making and low-level motor control through a \textbf{differentiable latent intent bottleneck}. 
    \textbf{(Left)} 
    System-2 (VLM) performs latent world modeling (LWM) to synthesize visual foresight within its native ViT feature space. This foresight serves as a structural bottleneck to convey the VLM's intent, which System-1 (Policy) then decodes into actions via latent inverse dynamics.
    A decoupled-to-unified training paradigm ensures stability, leveraging initial alignment in a consistent latent space to facilitate subsequent end-to-end refinement via action-aware gradients. 
    % System-2 (VLM) uses latent world modeling (LWM) to synthesize visual foresight as an intent bottleneck, decoded into actions by System-1 (Policy) via latent inverse dynamics. Stability is maintained through a two-stage training paradigm, from initial latent alignment in a consistent feature space to end-to-end action-aware refinement. 
    \textbf{(Right)}
    Powered by this structural grounding, DIAL scales across heterogeneous human and robot data to achieve SOTA performance with high data efficiency. In real-world deployments, it demonstrates robust zero-shot generalization across diverse configurations, mastering tasks ranging from semantic object manipulation to complex multi-stage coordination.
    % Powered by this structural grounding, DIAL scales across heterogeneous human-robot data, achieving SOTA performance with $10\times$ higher data efficiency and robust zero-shot generalization to unseen real-world configurations. 
}
    \label{fig:teaser}
\end{figure}

\begin{abstract}
The development of Vision-Language-Action (VLA) models has been significantly accelerated by pre-trained Vision-Language Models (VLMs). However, most existing end-to-end VLAs treat the VLM primarily as a multimodal encoder, directly mapping vision-language features to low-level actions. This paradigm underutilizes the VLM’s potential role in high-level decision making and introduces training instability, frequently causing degradation of its rich semantic representations.
To address these limitations, we introduce \textbf{DIAL} (\textbf{D}ecoupling \textbf{I}ntent and \textbf{A}ction via \textbf{L}atent World Modeling), a framework that bridges high-level decision making and low-level motor execution through a differentiable latent intent bottleneck. Specifically, a VLM-based System-2 performs latent world modeling by synthesizing the latent visual foresight 
within the native feature space of the VLM’s vision encoder; this foresight explicitly encodes the VLM’s intent and serves as the structural bottleneck. A lightweight System-1 policy then decodes this predicted intent together with the current observation into precise robot actions via latent inverse dynamics.
To ensure optimization stability, we employ a two-stage training paradigm: a decoupled warmup phase in which System-2 learns to predict latent futures while System-1 learns motor control under ground-truth future guidance  within a unified feature space, followed by seamless end-to-end joint optimization. This design enables action-aware gradients to refine the VLM backbone in a controlled manner while preserving its pre-trained knowledge.
Extensive experiments on the RoboCasa GR1 Tabletop benchmark demonstrate that DIAL establishes a new state of the art, achieving superior performance with 10$\times$ fewer demonstrations than prior methods. 
Furthermore, by leveraging heterogeneous human demonstrations, DIAL learns physically grounded manipulation priors and exhibits robust zero-shot generalization to unseen objects and novel configurations during real-world deployment on a humanoid robot.
\end{abstract}

\section{Introduction}

The development of generalist embodied agents has been significantly accelerated by pre-trained Vision-Language Models (VLMs)\cite{beyer2024paligemmaversatile3bvlm, chen2025eagle25boostinglongcontext, Qwen2.5-VL, Qwen3-VL}. By internalizing massive semantic knowledge from the internet, VLMs provide a unified cognitive foundation capable of handling diverse multimodal tasks. Consequently, using these pre-trained models as cognitive backbones for robot policies has become a dominant trend~\cite{rt22023arxiv,driess2023palmeembodiedmultimodallanguage,kim2024openvlaopensourcevisionlanguageactionmodel}. However, effectively translating the abstract, high-level intent of a VLM into high-frequency, precise motor control remains a major challenge.

\begin{figure}[!t]
    \centering
    \includegraphics[width=1.0\textwidth]{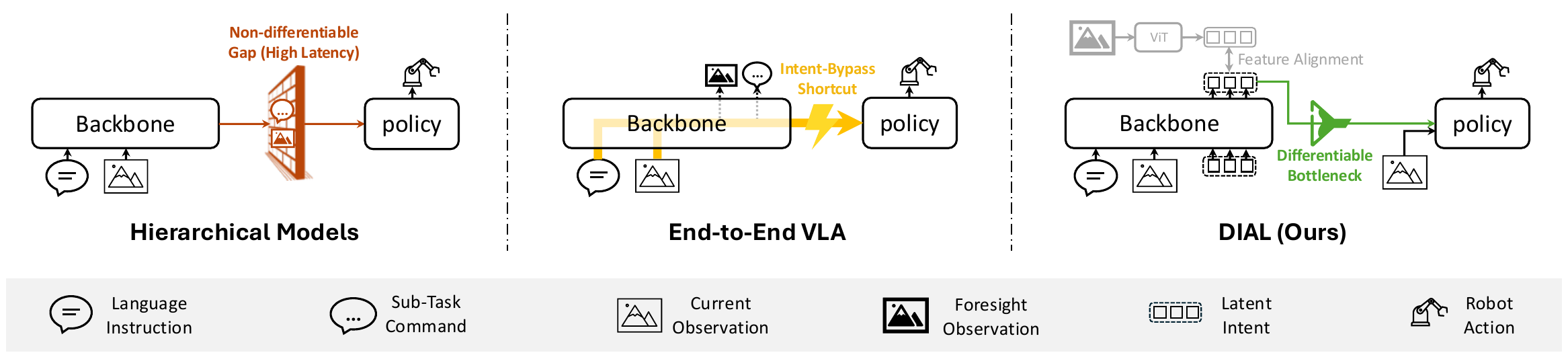}
    \caption{
    \textbf{Comparison of VLA Architectures. (Left) Hierarchical Models} decouple reasoning and execution via text or pixels, resulting in non-differentiable gaps and significant deployment latency. \textbf{(Middle) End-to-End VLAs} map multimodal features directly to actions.  
    Even when auxiliary tasks are used, they
    are typically treated as optional context, which cannot strictly guarantee that actions are grounded in
    the VLM’s intent. 
    \textbf{(Right) DIAL (Ours)} introduces a differentiable latent bottleneck. By requiring System-1 to bridge the gap between current visual features and System-2’s predicted latent foresight, DIAL ensures that execution is inherently anchored to the VLM’s predictive intent.
    }
    \label{fig:related_work}
\end{figure}

As summarized in Figure~\ref{fig:related_work}, 
existing approaches that leverage VLMs for robotic action generation face critical limitations. 
\textbf{Hierarchical Planners}~\cite{saycan2022arxiv, shi2025hirobotopenendedinstruction, huang2024rekepspatiotemporalreasoningrelational}
prompt the foundation model to generate high-level plans, typically text subtasks or code, to guide a separate low-level controller. While being interpretable and generalizable, this creates a non-differentiable wall that incurs high latency and prevents downstream action gradients from refining the VLM's physical understanding. 
In contrast, \textbf{End-to-End VLAs}~\cite{rt22023arxiv,gr00tn1_2025,intelligence2025pi05visionlanguageactionmodelopenworld} directly predict continuous actions. In practice, however, these approaches often treat the VLM primarily as a large multimodal encoder that extracts vision-language features, rather than allowing it to  serve as a high-level decision maker that explicitly represents task intent. As a result, training under low-level action supervision can become unstable, frequently causing the VLM’s semantic representations to collapse and overfit to spurious action patterns. Although auxiliary world modeling objectives~\cite{zheng2025flare, Zhao_2025_CVPR, tian2024predictive} help by instilling physical dynamics and foresight, the absence of a strict structural bottleneck still permits the policy to rely on superficial correlations rather than truly translating the VLM’s intent into precise motor commands.

This leaves the field in a structural dilemma: \textit{How can we design an end-to-end VLA that strictly grounds the policy in the VLM's intent, seamlessly unifying cognitive generalization with execution precision?}

To resolve this dilemma, we present \textbf{DIAL} (\textbf{D}ecoupling \textbf{I}ntent and \textbf{A}ction via \textbf{L}atent-world-modeling), as shown in Figure~\ref{fig:teaser}. Inspired by the cognitive distinction between deliberate reasoning (System-2) and reflexive motor control (System-1), DIAL introduces \textit{latent visual foresight} as a fully differentiable structural bottleneck between reasoning and execution. Rather than generating text or raw pixels, DIAL tasks the VLM (System-2) with predicting the future subgoal state entirely within the native feature space of its Vision Transformer (ViT) encoder, elevating it from a passive feature encoder to an active decision maker whose goal-directed foresight explicitly encodes the VLM's \textit{latent intent} and structurally governs downstream execution. A lightweight flow-matching policy (System-1) then operates as a latent inverse dynamics model: it compares the current visual observation against this predicted intent to deduce the precise high-frequency motor commands needed to reach the anticipated goal.

This architectural decoupling yields three key advantages. First, it naturally supports a \textbf{stable decoupled warmup}: the VLM learns physical dynamics from diverse, action-free data while the policy independently masters sensorimotor control under ground-truth future guidance, preventing the gradient interference and representation collapse typical of naive joint training. Second, because the latent intent is continuous and provides a consistent interface between both systems, DIAL transitions smoothly into \textbf{end-to-end synergy}: action gradients flow back through the latent intent into the VLM, regularized by the same foresight reconstruction loss, encouraging the predicted intent to evolve into an actively "action-aware" representation without disrupting the VLM's pretrained knowledge. Third, DIAL enforces strict \textbf{structural grounding}. Unlike prior works that loosely append future features as auxiliary context, our inverse-dynamics design imposes a hard bottleneck: System-1 must resolve the discrepancy between current and predicted latent states to generate actions, effectively mitigating the shortcut learning that commonly afflicts existing VLAs.

We conduct extensive experiments to validate DIAL across both comprehensive simulations and real-world deployments. Latent visualizations further confirm that DIAL successfully grounds abstract linguistic instructions into a coherent, structurally aligned ``visual roadmap'' for the policy. In summary, our core contributions are:
\begin{itemize}
    \item \textbf{Novel VLA Architecture:} We propose {DIAL}, an end-to-end framework that structurally bridges the cognitive generalization of VLMs with the execution precision of low-level policies. By utilizing latent visual foresight as a differentiable bottleneck, we ensure the generated actions are strictly grounded in the VLM's reasoning intent.
    \item \textbf{Decoupled-to-Unified Training Paradigm:} 
    To ensure stable optimization, we introduce a targeted dual-warmup strategy. Using action-free data, the VLM warmup shift its abstract semantic knowledge toward physical world dynamics. Simultaneously, the policy independently learns to map low-level perception and specific visual goals into precise motor actions. This distinct dual-initialization prevents representation collapse and paves the way for seamless end-to-end fine-tuning.
    \item \textbf{State-of-the-Art Performance \& Scalability:} 
    DIAL achieves the highest reported performance on the RoboCasa GR1 Tabletop benchmark 
    while using only $10\%$ of robot demonstrations required by previous methods. 
    It further shows strong scalability by successfully absorbing knowledge from diverse, cross-embodiment human data. Deployments on the IRON-R01-1.11 humanoid robot validate its reliable physical execution and impressive zero-shot transfer to novel scenarios.  
\end{itemize}

\section{Related Work}

The integration of large pretrained foundation models~\cite{beyer2024paligemmaversatile3bvlm, chen2025eagle25boostinglongcontext, Qwen2.5-VL, Qwen3-VL, touvron2023llama2openfoundation, liu2023visualinstructiontuning} has driven a shift from task-specific policies to generalist robotic agents. By leveraging internet-scale pre-training, VLMs and Large Language Models (LLMs) provide embodied agents with robust semantic reasoning and instruction-following capabilities. Existing applications of these models in robotics can be broadly categorized into two dominant paradigms: hierarchical frameworks and end-to-end architectures. Within the latter, integrating predictive world modeling has recently emerged as a critical frontier to enhance physical grounding.

\textbf{Hierarchical Abstraction vs. Low-Level Control.} 
Hierarchical frameworks decouple high-level reasoning from low-level execution. Typically, LLMs or VLMs act as semantic planners, generating textual subtasks~\cite{saycan2022arxiv, shi2025hirobotopenendedinstruction, chen2024egoplanbenchbenchmarkingmultimodallarge} or executable codes~\cite{liang2023codepolicieslanguagemodel, geminiroboticsteam2025geminirobotics15pushing} to drive separate downstream policies. Alternatively, some approaches employ heavy video diffusion models to predict pixel-level goal images, followed by an inverse dynamics model to infer actions from observational histories~\cite{du2023learninguniversalpoliciestextguided}. However, these paradigms face critical challenges. Text-driven methods rely heavily on rigid human annotations and struggle with complex, high-frequency control due to deployment latency. Meanwhile, video-generation models incur prohibitive inference costs and lack the rich, internalized common-sense semantics inherent to VLMs. Fundamentally, the non-differentiable interface between the foundation model and the execution policy obstructs seamless collaboration, preventing the foundation model from acquiring the action-aware dynamics necessary for fine-grained manipulation.

\textbf{End-to-End Vision-Language-Action Models.} 
To bridge the gap between semantics and control, end-to-end VLA architectures directly map multimodal inputs to continuous robot actions. Early VLAs~\cite{rt22023arxiv,driess2023palmeembodiedmultimodallanguage,kim2024openvlaopensourcevisionlanguageactionmodel}) cast actions as discrete text tokens within the VLM's vocabulary, a paradigm further optimized by efficient action tokenization~\cite{pertsch2025fast} and parallel decoding strategies~\cite{kim2025fine}. Recently, a ``dual-system'' architecture has emerged as the dominant trend, pairing a VLM (System-2 for semantic understanding) with a lightweight, continuous action expert like diffusion or flow-matching based transformers (System-1 for precise execution)~\cite{ black2024pi_0, intelligence2025pi05visionlanguageactionmodelopenworld, 
driess2025knowledgeinsulatingvisionlanguageactionmodels,
gr00tn1_2025, gear2025gr00tn16, li2024cogact, cheang2025gr3technicalreport, 
zhai2025ignitingvlmsembodiedspace}. 
Despite empirical successes, an essential dilemma persists: end-to-end training frequently induces model collapse, degrading the VLM's pretrained knowledge. To mitigate catastrophic forgetting, current methods often truncate gradient flows~\cite{intelligence2025pi05visionlanguageactionmodelopenworld} or freeze the VLM backbone entirely~\cite{gr00tn1_2025}. Consequently, the VLM is relegated to a passive representation encoder, leaving its core decision-making potential underexplored. While some works introduce Embodied Chain-of-Thought (CoT)~\cite{zawalski2024robotic, lee2025molmoactactionreasoningmodels} to reactivate reasoning, these methods require costly annotations and introduce severe inference latency.

\textbf{World Modeling within End-to-End VLAs.} 
To address the lack of physical foresight in reactive policies and the weak coupling between high-level semantic reasoning and low-level action execution, recent work has increasingly integrated world modeling objectives.
By anticipating future states, models are forced to develop a grounded understanding of physical dynamics. Early attempts~\cite{wu2024unleashing, cheang2024gr2generativevideolanguageactionmodel} utilize video pre-training to initialize policies with temporal priors. SEER~\cite{tian2024predictive} extends this by incorporating foresight-related features as an explicit condition for action generation. 
A prominent direction pursues unified autoregressive frameworks that jointly model sequences of discrete goal-image and actions tokens~\cite{Zhao_2025_CVPR, wang2026unified, cen2025worldvlaautoregressiveactionworld}, enabling tighter integration of world prediction and decision-making. 
Rather than relying on expensive raw pixel-level generation, many recent methods achieve greater efficiency by predicting compact latent dynamics, such as through discrete latent actions or motion tokens that model inter-frame changes~\cite{ye2025latent, Chen_2025_ICCV, chen2026villax}.
Alternatively, predicting continuous latents offers a more scalable approach: FLARE~\cite{zheng2025flare} leverages additional query tokens to align intermediate features with visual foresight, while UniCoD~\cite{zhang2025unicodenhancingrobotpolicy} employs a Mixture-of-Transformers (MoT) to unify world prediction and action execution.
However, a fundamental limitation remains: most existing architectures treat visual foresight either as an auxiliary regularization task or simply append it to a long-context sequence. 
This loose coupling may be insufficient to enforce a strict causal dependency between the anticipated world states and the execution policy. Consequently, there remains a risk that models circumvent true physical understanding, degenerating instead into spurious shortcut learning.

\section{Methodology}
\begin{figure}[!t]
    \centering
    \includegraphics[width=1.0\textwidth]{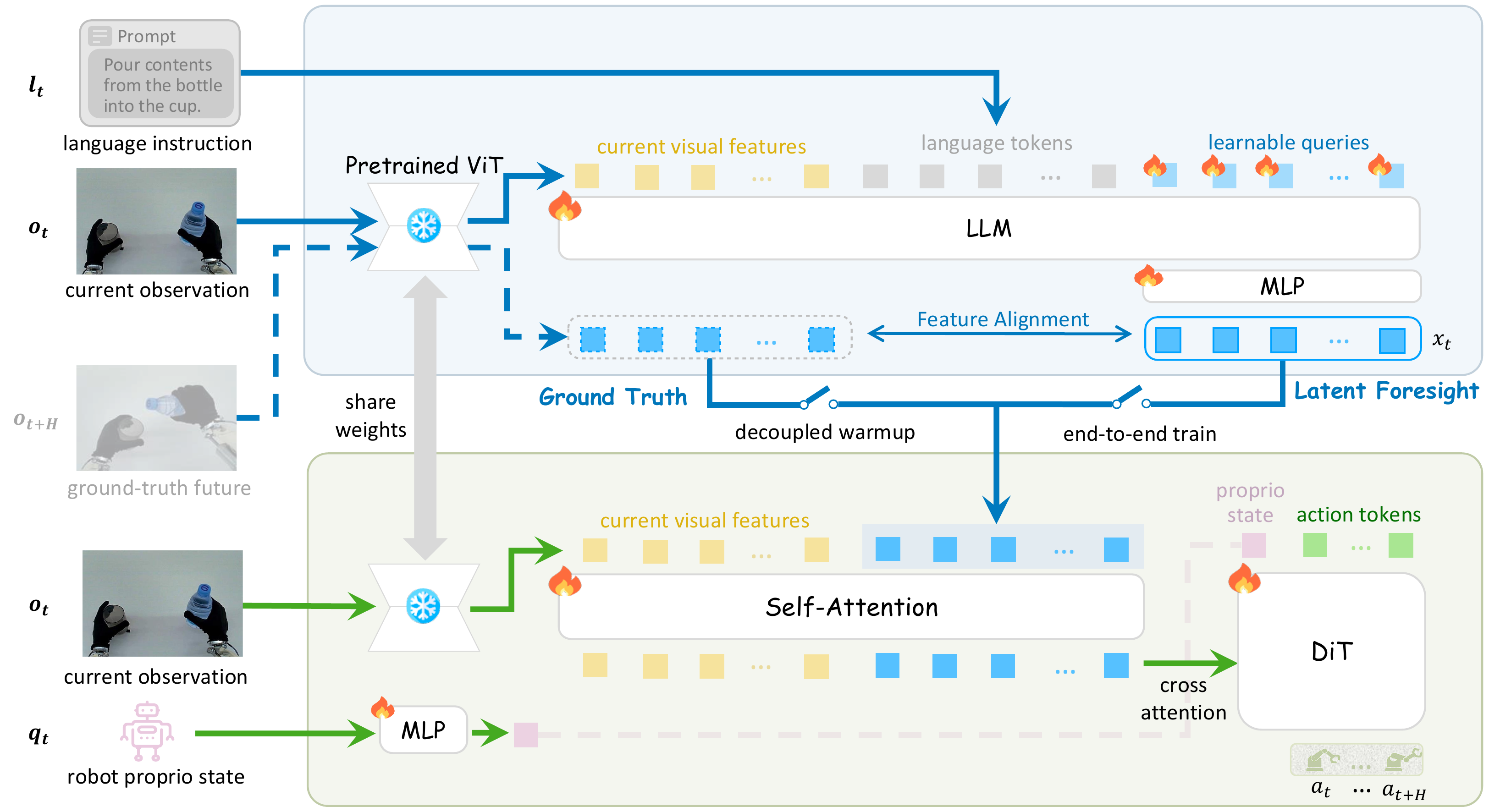}
    \caption{
    \textbf{The Dual-System Architecture of DIAL.}
    Built upon a {pre-trained VLM}, \textbf{System-2} (top) synthesizes a latent foresight ($x_t$) from language ($l_t$), current visual observation ($o_t$), and learnable queries via its LLM backbone and an MLP head. 
    \textbf{System-1} (bottom) employs self-attention to fuse current and foresight visual features, serving as the cross-attention condition for a DiT-based action decoder. This decoder directly takes the projected proprioceptive state ($q_t$) and noisy action tokens to generate action chunks. 
    To ensure feature consistency, both systems share the VLM's frozen pre-trained ViT. 
    As indicated by the switches, the training transitions from a \textbf{decoupled warmup} (conditioned on ground-truth features of $o_{t+H}$) to \textbf{end-to-end optimization} (conditioned on $x_t$). 
    Throughout both stages, an MSE loss is applied to align the latent foresight with ground-truth features.
    }
    \label{fig:model_overview}
\end{figure}

\subsection{Model Overview}

DIAL adopts a biologically inspired dual-system architecture to decouple high-level cognitive reasoning from low-level reactive execution, as illustrated in Figure~\ref{fig:model_overview}. At each timestep $t$, given a language instruction $l_t$, the current visual observation $o_t$, and the robot's proprioceptive state $q_t$, DIAL generates an action chunk $A_t =[a_t, a_{t+1}, \dots, a_{t+H-1}]$ with a horizon $H$. The overall policy is formulated as a sequential composition of predictive intent generation and reflexive action decoding:

\begin{equation}
    x_t = f_{\text{System-2}}(l_t, o_t), \quad A_t \sim \pi_{\text{System-1}}(\cdot \mid x_t, o_t, q_t)
\label{eq:dial}
\end{equation}

Specifically, \textbf{System-2} (analogous to the ``Brain'') leverages a pretrained VLM to process multimodal inputs and synthesize a \textbf{latent intent} $x_t$. Instead of directly outputting discrete actions, System-2 is tasked with \textit{latent world modeling} to envision the visual foresight of a future subgoal. Subsequently, \textbf{System-1} (the ``Cerebellum'') operates as a flow-matching-based reactive policy. It compares the current observation $o_t$ against the predictive intent $x_t$, grounding the high-level semantic goal into precise, high-frequency motor commands.

\subsection{System-2: Predictive Intent Synthesis via Latent World Modeling}

System-2 utilizes a pre-trained VLM (e.g., Qwen2.5-VL-3B~\cite{Qwen2.5-VL}) as its cognitive backbone. To endow the model with spatially aware predictive capabilities without the computational overhead of pixel-level reconstruction, we append $N$ learnable query tokens to the LLM's input sequence. The LLM processes these tokens alongside the visual patches of $o_t$ and the instruction $l_t$. The output representations corresponding to these queries are then passed through an {MLP projection head} to synthesize the latent intent $x_t \in \mathbb{R}^{N \times d}$. The number of queries $N$ is set to match the number of visual patches extracted by the ViT from a single observation, preserving spatial structure.

We explicitly constrain $x_t$ to encapsulate {visual foresight} by aligning it with the future state of the environment. Specifically, $x_t$ is trained to predict the visual representation of the observation $o_{t+H}$ at $H$ timesteps ahead. To ensure optimization stability and strictly align feature spaces, the target foresight representation is extracted using the \textit{identical} pre-trained ViT encoder, $\text{Enc}_{\text{ViT}}(\cdot)$, shared with the VLM backbone. The latent world modeling objective is optimized via Mean Squared Error (MSE):

\begin{equation}
    \mathcal{L}_{\text{world}} = \left\| x_t - \text{Enc}_{\text{ViT}}(o_{t+H}) \right\|^2_2
\label{eq:mse_loss}
\end{equation}

By minimizing this loss, System-2 learns to translate abstract semantic instructions into a structured, forward-looking latent representation, providing actionable predictive guidance for System-1.

\subsection{System-1: Reactive Motor Control as Latent Inverse Dynamics}

Given the predictive intent $x_t$ from System-2, {System-1} focuses purely on resolving the immediate physical requirements to reach that anticipated future. 

To achieve this, System-1 utilizes an independent perceptual pathway powered by the \textit{same} pre-trained ViT encoder shared with System-2. This architectural choice enforces strict feature-space consistency. By mapping both the current observation $o_t$ and the future intent $x_t$ into a unified latent manifold, System-1 can directly discern fine-grained spatial and dynamic discrepancies without cross-modal alignment overhead.

Architecturally, System-1 employs a lightweight 4-layer {self-attention module} to fully fuse the multi-modal visual context, taking the current visual features $\text{Enc}_{\text{ViT}}(o_t)$ and the predictive intent $x_t$ as inputs. The resulting spatially-aware fused representation serves as the conditioning signal for a 16-layer Diffusion Transformer (DiT), integrated via {cross-attention layers}. Concurrently, the low-dimensional proprioceptive state $q_t$ is projected into a dense feature token via an {MLP} and fed {directly} into the DiT as part of the input sequence, alongside the noisy action tokens. 

We formulate the action generation as an optimal transport problem using \textit{flow matching}. Given a ground-truth action chunk $A_t =[a_t, a_{t+1}, \dots, a_{t+H-1}]$, a time variable $\tau \sim \mathcal{U}[0, 1]$, and Gaussian noise $\epsilon \sim \mathcal{N}(\mathbf{0}, \mathbf{I})$, we define the interpolated path $A_t^\tau = \tau A_t + (1-\tau)\epsilon$. System-1 learns a velocity field $V_\theta$ to approximate the target vector field by minimizing:

\begin{equation}
    \mathcal{L}_{\text{fm}}(\theta) = \mathbb{E}_{\tau, \epsilon} \left[ \left\| V_\theta(A_t^\tau \mid x_t, \text{Enc}_{\text{ViT}}(o_t), q_t, \tau) - (A_t - \epsilon) \right\|^2_2 \right]
\label{eq:fm_loss}
\end{equation}

Conceptually, System-1 functions as a {latent inverse dynamics model}. Unlike traditional inverse models that compute actions directly from high-dimensional, noisy raw pixels, our approach resolves state-transition dynamics entirely within a structured latent space. 
This hierarchical separation isolates low-level execution from high-level reasoning, ensuring robust and high-frequency motor control.

\subsection{Optimization Strategy: From Decoupled Warmup to End-to-End Synergy}

The hierarchical design of DIAL naturally supports a stable two-stage training paradigm, transitioning from independent module initialization to fully differentiable joint optimization. Throughout both stages, all parameters of System-1 remain fully trainable. For {System-2}, we freeze the pre-trained ViT encoder and the text embedding layer of the VLM, while the rest of the network (including the LLM blocks, learnable queries, and the MLP projection head) is fully updated.

\textbf{Stage 1: Decoupled Warmup.} We first pre-train System-1 and System-2 independently to prevent posterior collapse. System-2 is optimized solely via $\mathcal{L}_{\text{world}}$ (Eq.~\ref{eq:mse_loss}) to master physically-grounded semantic reasoning and visual foresight. Concurrently, System-1 is trained via $\mathcal{L}_{\text{fm}}$ (Eq.~\ref{eq:fm_loss}) by substituting $x_t$ with the \textit{ground-truth} future visual features $\text{Enc}_{\text{ViT}}(o_{t+H})$. This ensures the ``Cerebellum'' learns optimal sensorimotor control under perfect future guidance, while the ``Brain'' concurrently learns to imagine that future.

\textbf{Stage 2: End-to-End Training.} Following the warmup, we unify the pipeline. System-1 is now conditioned directly on the synthesized latent intent $x_t$ generated by System-2. Crucially, unlike traditional inverse dynamics pipelines that are disjointed and difficult to optimize jointly, our latent formulation is \textit{fully differentiable}. This enables seamless end-to-end training, allowing downstream action-generation gradients (via $\mathcal{L}_{\text{fm}}$) to backpropagate smoothly through $x_t$ and directly into the trainable parameters of the VLM backbone. 

This gradient feedback is the cornerstone of DIAL: it forces the intent $x_t$ to become explicitly {action-aware}. Rather than remaining a pure visual prediction, $x_t$ evolves into a task-oriented representation strictly optimized for downstream motor execution. The overall training objective seamlessly combines both losses:

\begin{equation}
    \mathcal{L}_{\text{total}} = \mathcal{L}_{\text{world}} + \mathcal{L}_{\text{fm}}
\label{eq:total_loss}
\end{equation}

This joint optimization effectively marries high-level cognitive planning with low-level physical grounding, providing a highly scalable template for end-to-end Vision-Language-Action (VLA) models.

\section{Experimental Setup}

\subsection{Benchmarks and Datasets}

\subsubsection{RoboCasa GR1 Tabletop Simulation}

\begin{figure}[!t]
    \centering
    \includegraphics[width=1.0\textwidth]{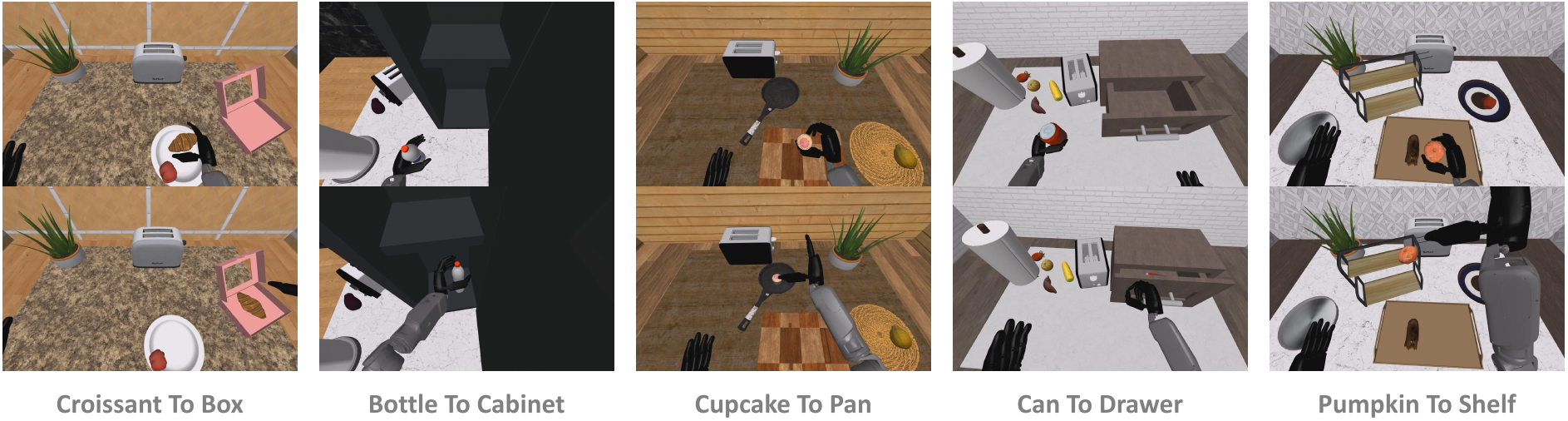}
    \caption{
    Examples from the 24 RoboCasa GR1 Tabletop Tasks, including object rearrangement (e.g., Croissant to Box) and interaction with articulated fixtures (e.g., Bottle to Cabinet).
    }
    \label{fig:robocasa_case}
\end{figure}

We conduct simulation experiments on the RoboCasa benchmark utilizing the GR1 robot. As illustrated in Figure~\ref{fig:robocasa_case}, our evaluation suite comprises 24 tabletop tasks, each assessed over 50 episodes. This suite includes 18 ``Pick-and-Place'' rearrangement tasks, where the robot follows language instructions to move objects between containers, and 6 ``Articulated tasks'' that involve more complex interactions such as placing objects inside and subsequently closing cabinets, drawers, or microwaves. 
We represent the robot's state and actions using a 47-dimensional vector. This vector comprises 29 joint-space degrees of freedom (DoF)—including the dual arms (14), hands (12), and waist (3)—along with 18 dimensions for the end-effector (EEF) poses (3D position and 6D rotation for each wrist). While the robot is ultimately commanded via its 29-dimensional joint space, we include the EEF poses to align the robot’s representation with human data, following standard practice~\cite{cheang2025gr3technicalreport}. Across all simulation tasks, visual observations are processed at a $224 \times 224$ resolution, utilizing $N=64$ learnable queries for latent intent synthesis, and actions are predicted with a chunk size of $H=16$.

\paragraph{Robot-Only Training Regimes.} To evaluate effectiveness and data efficiency, we assess our method under two robot-only settings. The \textbf{Full Data} regime (Figure~\ref{fig:results_robocasa_full}) utilizes 24,000 trajectories (1,000 per task) and involves 160,000 training steps. The \textbf{Few-Shot} regime (Figure~\ref{fig:results_robocasa_small}) utilizes a 10\% subset of 2,400 trajectories (100 per task) trained for 40,000 steps. For DIAL, both regimes follow a two-stage training schedule: the first half of the steps are dedicated to decoupled warmup, followed by end-to-end training for the remainder.

\paragraph{Learning from Human Data.} 
To further examine the scalability of DIAL, we investigate its ability to leverage large-scale human demonstrations to enhance generalization (Figure~\ref{fig:results_robocasa_small}). We incorporate the \texttt{basic\_pick\_place} subset of the EgoDex dataset~\cite{hoque2025egodex}, which contains 27,419 trajectories of non-articulated pick-and-place interactions. To align human annotations with the robot's 47-dimensional state space, we extract the wrist EEF poses—the shared components between both embodiments—and pad the remaining dimensions. For human data, the state at time $t+1$ is treated as the ground-truth action for the state at $t$. This large-scale dataset is combined with the few-shot robot set for a two-stage training process: a 40,000-step pre-training phase (split evenly between warmup and end-to-end training), followed by 20,000 steps of end-to-end fine-tuning exclusively on the few-shot robot data.

\paragraph{Generalization Scenarios.} To rigorously verify the generalization gains from human-centric priors, we constructed three Out-of-Distribution (OOD) testing scenarios using the assets from the RoboCasa benchmark:  (1) \textit{Unseen Appearance (18 Tasks)}, which introduces novel visual textures to familiar source-target container pairs; (2) \textit{Unseen Combinations (14 Tasks)}, which requires manipulating seen objects within novel container pairings not encountered during training; and (3) \textit{Unseen Object Types (32 Tasks)}, which tests the model's ability to generalize to entirely novel object categories across 32 different container combinations.

\subsubsection{Real-World Experiments}

\begin{figure}[!htbp]
    \centering
    \includegraphics[width=1.0\textwidth]{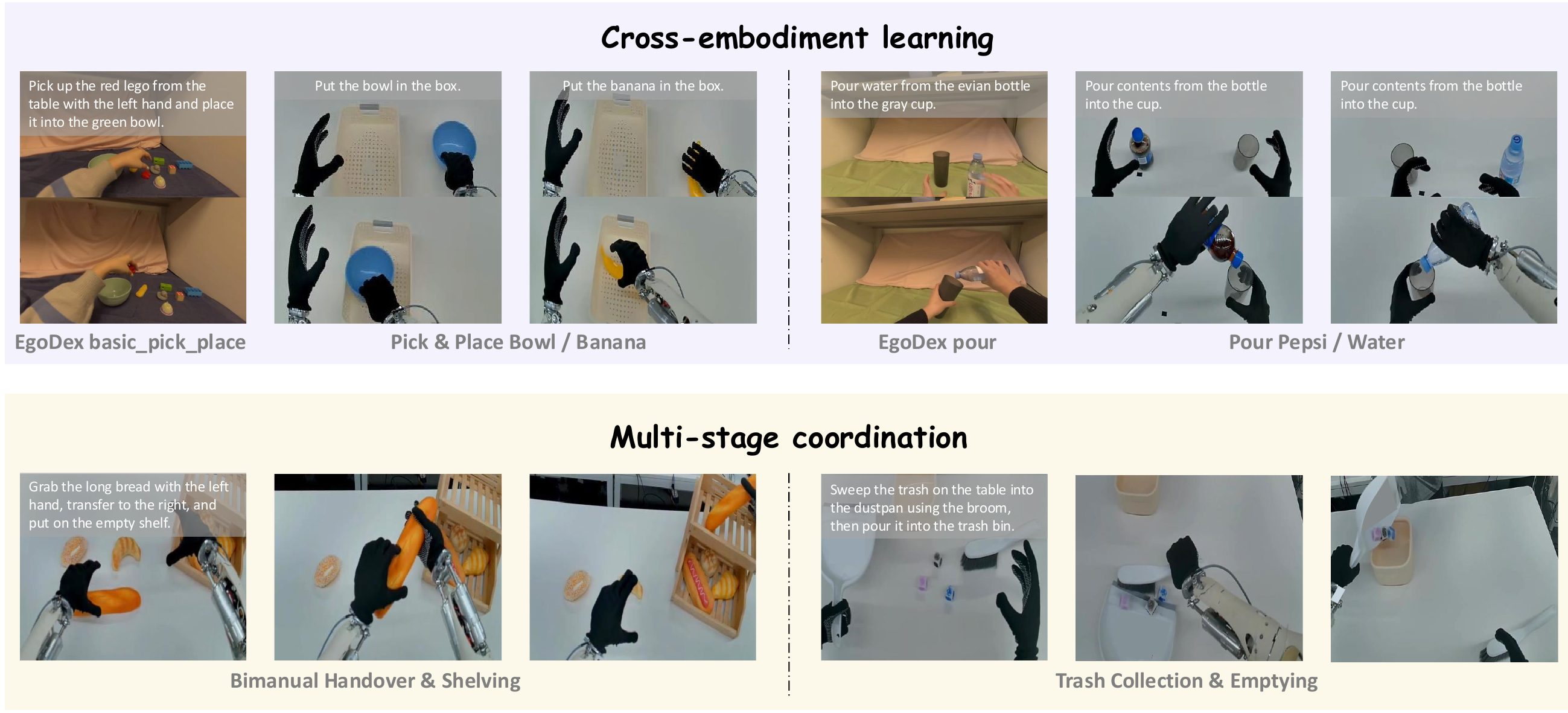}
    \caption{
    % \textbf{
    % Real-world Tasks and Data Sources.} Comparison between human demonstrations from the EgoDex dataset and corresponding robot executions (Pick \& Place and Pouring) used for cross-embodiment learning on the IRON-R01-1.11 robot.
    % }
    % \textbf{Real-world Tasks and Data Sources.} We evaluate our framework across a diverse suite of manipulation tasks categorized by data composition and task complexity: (Top) \textbf{Cross-embodiment learning}, featuring \textit{Pick \& Place} (relocation) and \textit{Pouring} (tilting), which are jointly trained on a mixture of EgoDex human demonstrations and corresponding robot trajectories. (Bottom) \textbf{Multi-stage coordination}, showing more complex skills such as \textit{Handover} (inter-hand object exchange) and \textbf{Sweeping} (tool-mediated cleaning). 
    % While the top tasks leverage heterogeneous data sources, the bottom tasks are trained on robot-native sequences to capture intricate synchronization and tool-use behaviors.
    \textbf{Real-world Tasks and Data Sources.} We evaluate our framework across manipulation tasks categorized by data composition and complexity: (Top) \textbf{Cross-embodiment learning}, featuring \textit{Pick \& Place} and \textit{Pouring}, jointly trained on EgoDex human and robot trajectories. (Bottom) \textbf{Multi-stage coordination}, including \textit{Handover} and \textit{Sweeping}. While top tasks leverage heterogeneous data, bottom tasks use robot-native sequences to capture intricate synchronization and tool-use.
    }
    \label{fig:demo_task}
\end{figure}

\begin{figure}[!htbp]
    \centering
    \includegraphics[width=1.0\textwidth]{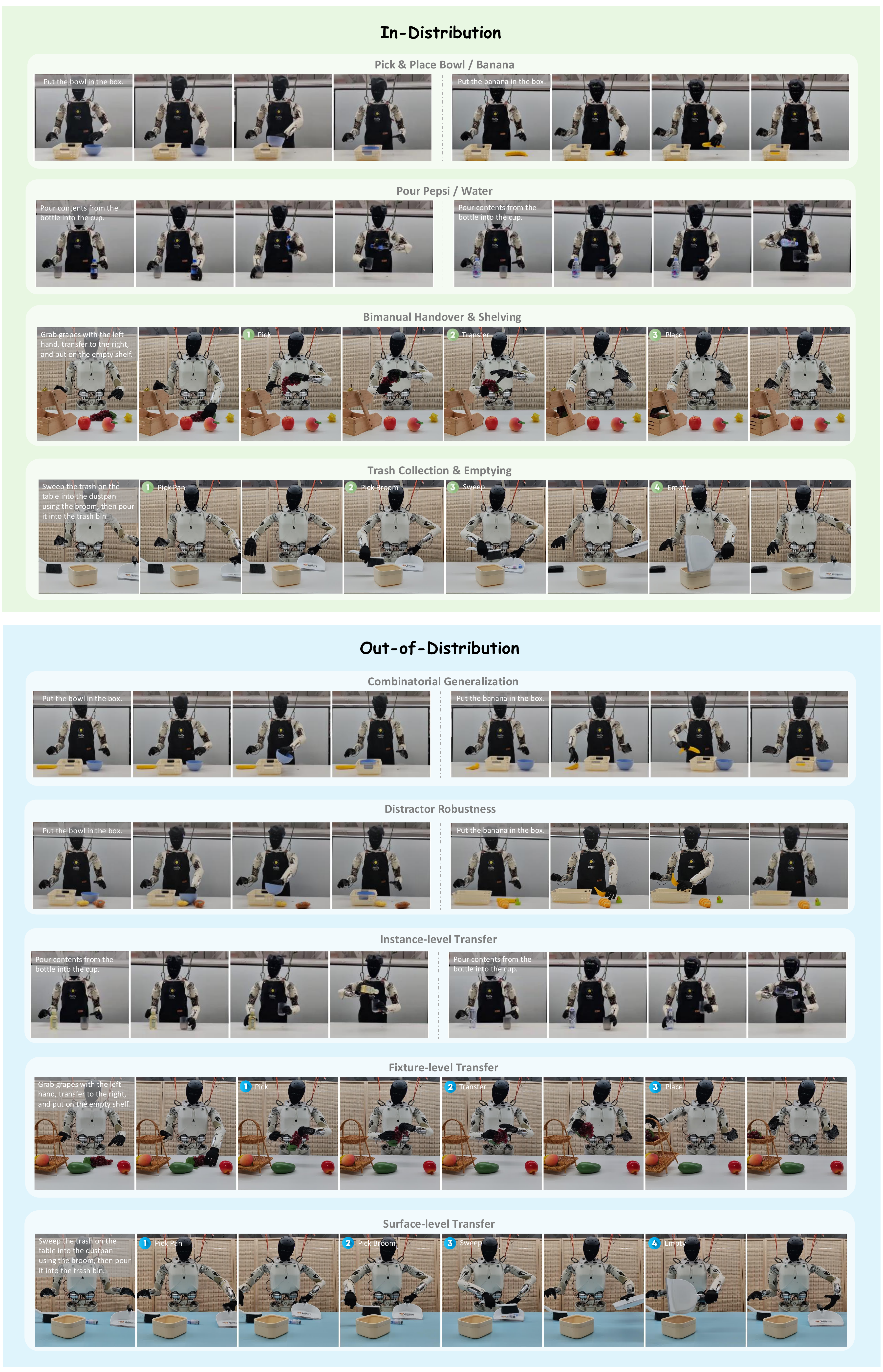}
    \caption{
    \textbf{Real-world Generalization Scenarios.} Comparison of in-distribution tasks and five OOD categories: combinatorial generalization (multiple seen objects), distractor robustness (unseen background items), instance-level transfer (novel object types), fixture-level transfer (novel shelf types), and surface-level transfer (unseen tablecloths).
    }
    \label{fig:OOD_task}
\end{figure}

We validate our method on the IRON-R01-1.11 robot using a 50-dimensional state and action space. 
This setup extends the simulation configuration (arms, hands, waist, and EEF poses) by incorporating an additional 3-DoF head.
% As shown in Figure~\ref{fig:demo_task}, 
% we design two real-world tasks analogous to representative EgoDex subsets:
% \begin{itemize}[leftmargin=*]
%     \item \textbf{Pick \& Place:} Mimicking the EgoDex \texttt{basic\_pick\_place} subset (27,419 trajectories), the robot must pick up an object (e.g., a bowl or banana) and place it into a box.
%     \item \textbf{Pouring:} Mimicking the EgoDex \texttt{pour} subset (3,205 trajectories), the robot must grasp a bottle with one hand and a cup with the other, performing a pouring motion.
% \end{itemize}
As shown in Figure~\ref{fig:demo_task}, we design four real-world tasks categorized into two regimes based on their data sources and coordination complexity:

\textbf{Cross-embodiment learning} (Jointly trained with EgoDex human demonstrations):
\begin{itemize}[leftmargin=*]
    \item \textbf{Pick \& Place:} Mimicking the EgoDex \textit{basic\_pick\_place} subset (27,419 trajectories), the robot must pick up an object (e.g., a bowl or banana) and place it into a box.
    \item \textbf{Pouring:} Mimicking the EgoDex \textit{pour} subset (3,205 trajectories), the robot must grasp a bottle with one hand and a cup with the other, performing a coordinated tilting motion.
\end{itemize}

\textbf{Multi-stage coordination} (Trained on robot-native sequences):
\begin{itemize}[leftmargin=*]
    \item \textbf{Handover \& Shelving:} 
    % The robot must grasp an object with the left hand, perform an inter-hand transfer to the right hand, and precisely place it onto a target shelf.
    A three-stage task requiring  inter-hand coordination: (i) picking the object with the left hand, (ii) transferring it to the right hand, and (iii) placing it onto the target shelf.
    \item \textbf{Trash Collection \& Emptying:} 
    % The robot must grasp a dustpan and a broom, sweep tabletop debris into the dustpan, and subsequently dispose of the contents into a trash bin.
    A four-stage task involving tool-mediated manipulation: (i) grasping the dustpan, (ii) grasping the broom, (iii) sweeping tabletop debris into the dustpan, and (iv) emptying the contents into a trash bin.
\end{itemize}

% \paragraph{Training Protocol.} 
% For each task, we collected 120 robot trajectories in a laboratory setting. All models follow a two-stage training paradigm: a large-scale pre-training phase of 160,000 steps on a mixture dataset comprising 32k proprietary factory-collected robot trajectories and 30k EgoDex trajectories, followed by a task-specific fine-tuning phase of 2,000 steps. To ensure a fair comparison, DIAL allocates the pre-training phase into 80,000 steps of decoupled warmup and 80,000 steps of end-to-end training, matching the total training steps of the baselines.

\textbf{Training Protocol.} For each real-world task, we collected 120 robot trajectories in a laboratory setting. To maintain architectural consistency, visual observations for the real-world robot are kept at a $224 \times 224$ resolution with $N=64$ queries, identical to the simulation setup. All models follow a two-stage training paradigm: a large-scale pre-training phase of 160,000 steps followed by a task-specific fine-tuning phase of 2,000 steps. 
% Across all experiments, visual observations are unified to a $224 \times 224$ resolution, which corresponds to $N=64$ queries for synthesizing the latent intent.

For \textbf{Cross-embodiment learning} tasks, we we retain the action chunk size of $H=16$ and pre-train on a mixture dataset comprising 32k proprietary factory-collected trajectories and 30k EgoDex trajectories. 
However, for \textbf{Multi-stage coordination} tasks, we employ a larger action chunk size of $H=50$ 
% to capture the extended temporal dependencies of complex bimanual behaviors; these models are pre-trained 
and pre-train exclusively on robot trajectories, as the EgoDex dataset lacks corresponding human demonstrations for such specific coordination behaviors. 
To ensure a fair comparison, DIAL allocates the pre-training phase into 80,000 steps of decoupled warmup and 80,000 steps of end-to-end training across all settings, matching the total training steps of the baselines.

\paragraph{Generalization Scenarios.} To further evaluate the model's robustness and generalization capabilities, we establish five OOD testing scenarios, as demonstrated in Figure~\ref{fig:OOD_task}. 
(1) \textit{Combinatorial Generalization}: In the pick-and-place task, both a banana and a bowl are simultaneously present in the workspace, requiring the model to correctly disambiguate and follow specific language instructions. 
(2) \textit{Distractor Robustness}: We introduce unseen objects into the workspace as distractors while the target object (e.g., a banana or bowl) is present, testing the policy's resilience to visual clutter. 
(3) \textit{Instance-level Transfer}: In the pouring task, the model is required to manipulate novel object instances, such as beverage or mineral water bottles with previously unseen geometries, sizes, or liquid colors.
(4) \textit{Fixture-level Transfer}:
In the handover task, the model must adapt to a previously unseen multi-tier shelf with a novel design.
% In the handover task, the model must adapt to visual and geometric variations of placement equipment, specifically performing precise deposits on previously unseen multi-tier shelf designs. 
(5) \textit{Surface-level Transfer}: 
In the sweeping task, the model is required to perform cleanup actions on a workspace covered with an unseen tablecloth.
% In the sweeping task, the model must maintain robust execution over unseen working surfaces, such as generalizing from a standard tabletop to a smooth, solid-colored tablecloth.

\subsection{Compared Methods}

We evaluate DIAL against a diverse set of methods, ranging from established policy learning frameworks to state-of-the-art VLA architectures, as well as several controlled variants designed to isolate component contributions.

\subsubsection{Representative Prior Arts}

We first compare DIAL with standard policy learning frameworks, including: \textbf{Diffusion Policy}~\cite{chi2023diffusion}, which models action generation via U-Net denoising; \textbf{UWM}~\cite{li2025unified}, a transformer unifying action and video diffusion processes; and \textbf{FLARE}~\cite{zheng2025flare}, a flow-matching framework with future latent alignment. For VLA comparisons, we include \textbf{GR00T-N1.6}~\cite{gear2025gr00tn16}, an upgraded GR00T framework incorporating a larger DiT backbone and fine-tuned late-stage VLM layers.

\noindent\textbf{Qwen3-based VLAs.} To ensure a fair comparison against models equipped with cutting-edge VLM backbones, we evaluate several architectures implemented via \textit{StarVLA}~\cite{starvla2025}, all utilizing {Qwen3-VL}~\cite{Qwen3-VL}:
\begin{itemize}[leftmargin=*]
    \item \textbf{GR00T-Qwen3}~\cite{gr00tn1_2025}: Adapts the GR00T dual-system, combining a frozen Qwen3-VL for vision-language representations with a flow-matching DiT for action generation.
    \item \textbf{$\pi$-Qwen3}~\cite{black2024pi_0}: Couples per-layer VLM representations with a flow-matching expert via KV caching.
    \item \textbf{FAST-Qwen3}~\cite{pertsch2025fast}: Employs FAST action tokenization for efficient autoregressive prediction.
    \item \textbf{OFT-Qwen3}~\cite{kim2025fine}: An optimized fine-tuning recipe featuring parallel action-chunk decoding.
\end{itemize}

\subsubsection{Controlled Variants and Ablations}

To rigorously isolate the impact of DIAL's specific architectural designs, we construct several controlled variants and ablation models, primarily based on the same {Qwen2.5-VL-3B}~\cite{Qwen2.5-VL} backbone used in DIAL:

\begin{itemize}[leftmargin=*]
    \item \textbf{GR00T-Qwen2.5 / -FT:} Replicates the GR00T architecture using Qwen2.5 as the VLM backbone. We consider two settings: (i) a frozen variant, where both the ViT encoder and the LLM are fixed, and (ii) an LLM-tuned variant (-FT), which updates the core language modeling blocks while keeping the ViT encoder and text embedding layers frozen.
    \item \textbf{GR00T-Qwen2.5 + FLARE / SEER:} Extensions of the fine-tuned (-FT) baseline that incorporate future-aligned query tokens, inspired by FLARE~\cite{zheng2025flare} and SEER~\cite{tian2024predictive}. Specifically, the FLARE-style variant uses these tokens solely for latent regularization, whereas the SEER-style variant explicitly concatenates them with vision-language features to condition the DiT.
    \item \textbf{GR00T-Qwen2.5 + SEER-EV:} Augments the SEER variant with an independent perceptual path (Extra Vision) for System-1. This isolates the effect of decoupled perception but lacks DIAL's specialized bottleneck design.
    \item \textbf{DIAL-DINO:} Replaces the visual foresight target (for System-2) and the current observation encoder (for System-1) with an external {DINO-v2}~\cite{oquab2024dinov2learningrobustvisual} encoder. Since System-2 still processes its current visual input via the VLM-native encoder, this variant tests the impact of forcing the high-level ``Brain'' and the low-level ``Body'' to communicate across mismatched latent spaces.
    \item \textbf{DIAL w/o Human Data:} Evaluates DIAL trained solely on robot data to quantify the gains from human data pre-training. Note that in certain evaluations where human data is entirely excluded for all baselines, the default ``DIAL'' entry represents this robot-only version. The explicit distinction is only made in Fig.\ref{fig:results_robocasa_egodex_narrow},  Fig.\ref{fig:results_real_robot_id}, and Fig.\ref{fig:results_real_robot_ood}.
\end{itemize}

\section{Experiments}
\label{sec:experiments}

In this section, we empirically evaluate DIAL. To structure our analysis, we formulate five core research questions:

\begin{itemize}
    \item \textbf{Q1 (Performance):} 
    Does DIAL outperform state-of-the-art baselines in task success rate and sample efficiency on the public humanoid simulation benchmark?
    \item \textbf{Q2 (Architecture):} 
    Which architectural designs are essential for grounding the VLM's high-level intent in low-level control?
    \item \textbf{Q3 (Scalability):} 
    Can DIAL effectively scale by leveraging heterogeneous human demonstrations to enhance both in-distribution and OOD performance?
    \item \textbf{Q4 (Robustness):} 
    % Can DIAL achieve robust real-world generalization, and what role does the decoupled warmup play in achieving this?
    Can DIAL exhibit robust real-world performance, and what role does the decoupled warmup play in stabilizing the deployment?
    \item \textbf{Q5 (Interpretability):} 
    Do the predicted latent foresights capture semantically meaningful, task-relevant dynamics in the VLM's native feature space?
\end{itemize}

% ----------------------------------------------------------------------
% Section 4.1: Addressing Q1
% ----------------------------------------------------------------------
\subsection{Overall Performance Comparison}
\label{sec:exp_sota}

\begin{figure}[!t]
    \centering
    \includegraphics[width=1.0\textwidth]{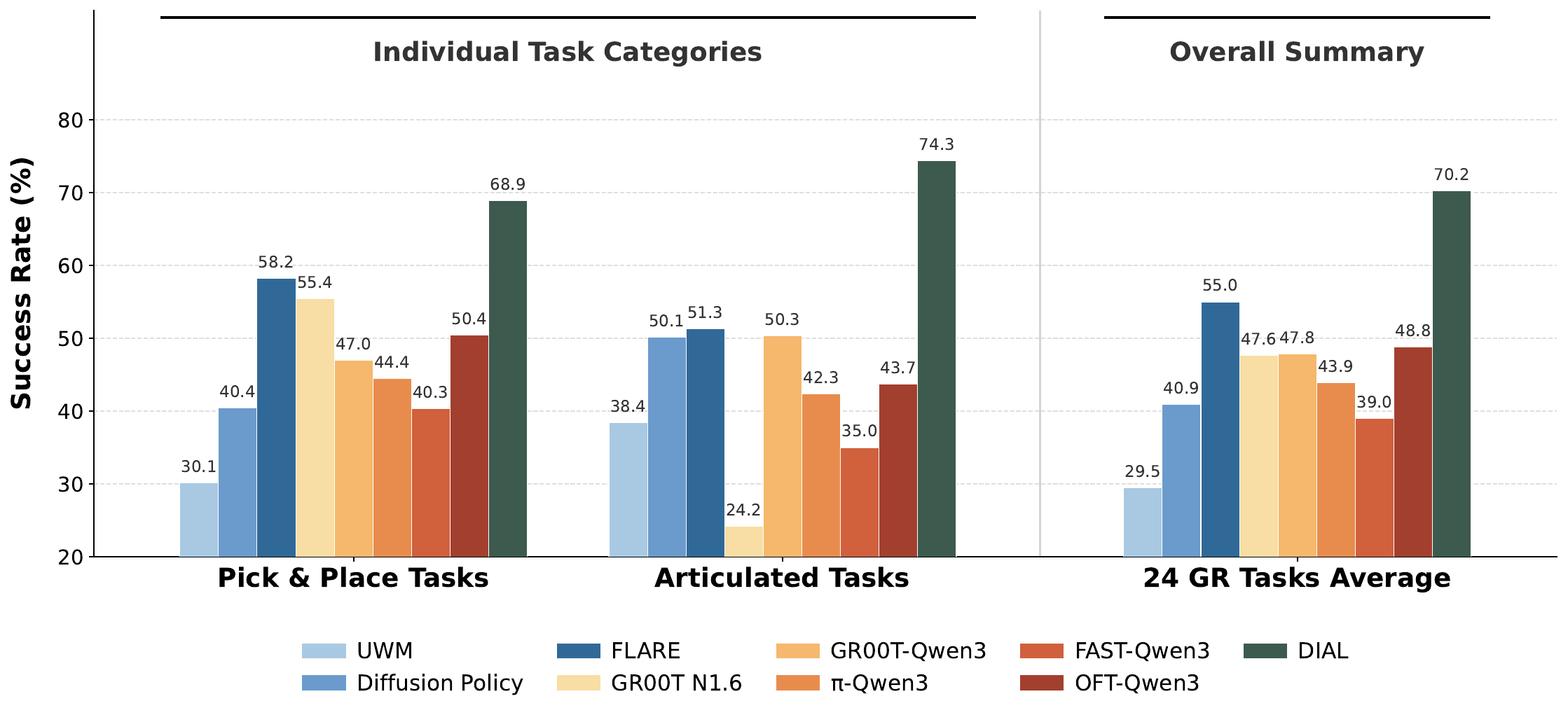}
    \caption{
    Results on RoboCasa GR1 Tabletop Simulation with full training data.
    }
    \label{fig:results_robocasa_full}
\end{figure}

We benchmark DIAL against state-of-the-art policies on the comprehensive RoboCasa GR1 Tabletop simulation suite. As shown in Figure~\ref{fig:results_robocasa_full}, DIAL achieves an average success rate of {70.2\%}, substantially outperforming the strongest baseline, FLARE (55.0\%), as well as advanced VLA architectures such as GR00T-N1.6 (47.6\%). This consistent margin establishes DIAL as the new state of the art on this benchmark.

A closer examination across task categories further reveals DIAL’s robustness. In {Pick \& Place} tasks, which require precise grounding of semantic instructions into object rearrangement behaviors, DIAL attains a {68.9\%} success rate, significantly surpassing existing VLA variants and demonstrating stronger instruction-to-action alignment. In {Articulated Tasks}, where the robot must manipulate articulated objects, DIAL achieves an even higher {74.3\%} success rate. The strong performance across both categories highlights DIAL’s balanced capability, maintaining consistently high effectiveness across distinct task types. We attribute this stability to DIAL’s dual-system decoupling, which provides a structured interface between high-level intent and low-level control.

To assess data efficiency, we further evaluate DIAL under a strict few-shot setting. As shown in Figure~\ref{fig:results_robocasa_small}, when trained with only {100 trajectories per task}, DIAL achieves a {58.3\%} success rate. Notably, this performance already surpasses FLARE (55.0\%) trained under the full-data regime with {1,000 trajectories per task} (Figure~\ref{fig:results_robocasa_full}). 
This {10$\times$ reduction} in demonstration requirements, while maintaining superior performance, highlights the strong inductive bias introduced by DIAL's structural bottleneck, enabling highly scalable and data-efficient robot learning.

% ----------------------------------------------------------------------
% Section 4.2: Addressing Q2
% ----------------------------------------------------------------------

\subsection{Ablation on Bridging Mechanisms}
\label{sec:exp_ablation}

\begin{figure}[!t]
    \centering
    \includegraphics[width=1.0\textwidth]{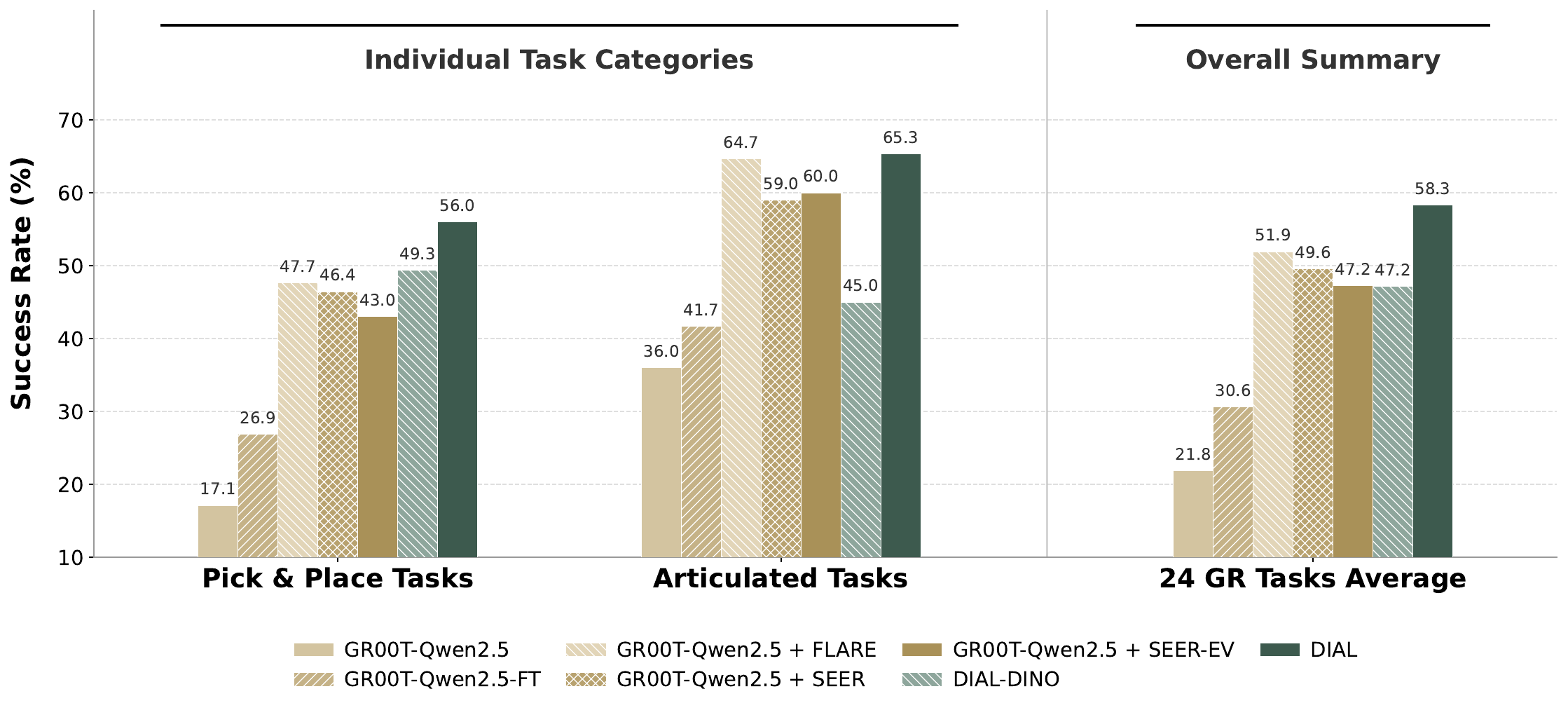}
    \caption{
    Results on RoboCasa GR1 Tabletop Simulation under the few-shot setting.
    }
    \label{fig:results_robocasa_small}
\end{figure}

To isolate the source of DIAL's improvements, we systematically decompose the architecture to evaluate the contributions of world modeling objectives, the System-1/System-2 interface design, and feature alignment. All ablations are conducted in the few-shot regime (100 trajectories per task), with results summarized in Figure~\ref{fig:results_robocasa_small}.

\paragraph{Effect of World Modeling Objectives.}
We begin with standard VLA baselines that primarily treat the VLM as a multimodal encoder without explicit world modeling supervision. The frozen {GR00T-Qwen2.5} baseline achieves only {21.8\%} success, and full fine-tuning ({-FT}) improves performance modestly to {30.6\%}. This limited gain suggests that merely updating parameters in a tightly coupled architecture is insufficient. Without an explicit objective that grounds latent representations in future world states, the policy tends to overfit to proprioceptive cues while under-utilizing complex visual observations. These results highlight the necessity of incorporating structured predictive signals rather than relying solely on scale or fine-tuning.

\paragraph{Interface Design.}
% : From Informational Shortcuts to Structural Bottlenecks.}
We next examine how intent should be communicated from System-2 to System-1. Architectures with \emph{loose coupling} treat the predicted intent as optional context. For instance, {+SEER} (49.6\%) concatenates future tokens to the policy input, while {+SEER-EV} (47.2\%) further provides System-1 with a direct visual perception path. Both variants fail to surpass the 50\% success rate. Similarly, {+FLARE}, which introduces future prediction as an auxiliary training objective without enforcing its usage at execution time, achieves {51.9\%}. 

When the intent signal is not structurally enforced, the policy tends to bypass the high-level intent and instead exploit superficial shortcuts, since there is no architectural constraint requiring the intent to be faithfully translated into action. Notably, adding an extra visual pathway in SEER-EV even degrades performance (49.6\% $\rightarrow$ 47.2\%), indicating that unconstrained access to raw features exacerbates the tendency to ignore cognitive foresight.

In contrast, {DIAL} introduces a \emph{structural bottleneck} by formulating System-1 as a latent inverse dynamics model, where the synthesized intent serves as an indispensable target rather than auxiliary context. This design enforces that actions must be derived by bridging the current observation and the explicitly predicted future state. As a result, DIAL achieves a state-of-the-art \textbf{58.3\%}, substantially outperforming all alternative interfaces in the low-data regime.

\paragraph{Feature Alignment.}
Finally, we investigate whether DIAL’s gains depend on the native VLM feature space. In {DIAL-DINO}, we replace the internal ViT features with DINO-v2 representations. Despite their strong geometric priors, performance drops from {58.3\%} to {47.2\%}. 

This degradation indicates that geometric richness alone is insufficient; what matters is latent consistency between reasoning and control. In DIAL-DINO, System-2 reasons in the VLM’s native semantic space but must project its intent into a different feature manifold for execution, creating a semantic--physical misalignment. By contrast, DIAL’s shared native ViT ensures that both systems operate within a unified latent space, eliminating cross-manifold translation and maximizing the fidelity of intent-to-action transfer.

Across all ablations, the results consistently demonstrate that effectively connecting a dual-system architecture requires more than auxiliary supervision or loose input concatenation. Superior performance is achieved only when (i) intent is grounded via explicit world modeling, (ii) the architecture enforces a structural dependency to prevent shortcut learning, and (iii) reasoning and control share a unified latent space. Together, these findings validate the necessity of DIAL’s core design: explicit system decoupling bridged by a strict, representationally consistent predictive bottleneck.

% ----------------------------------------------------------------------
% Section 4.3: Addressing Q3
% ----------------------------------------------------------------------

\subsection{Scalability via Human Data}
\label{sec:exp_human_data}

\begin{figure}[!t]
    \centering
    \includegraphics[width=\textwidth]{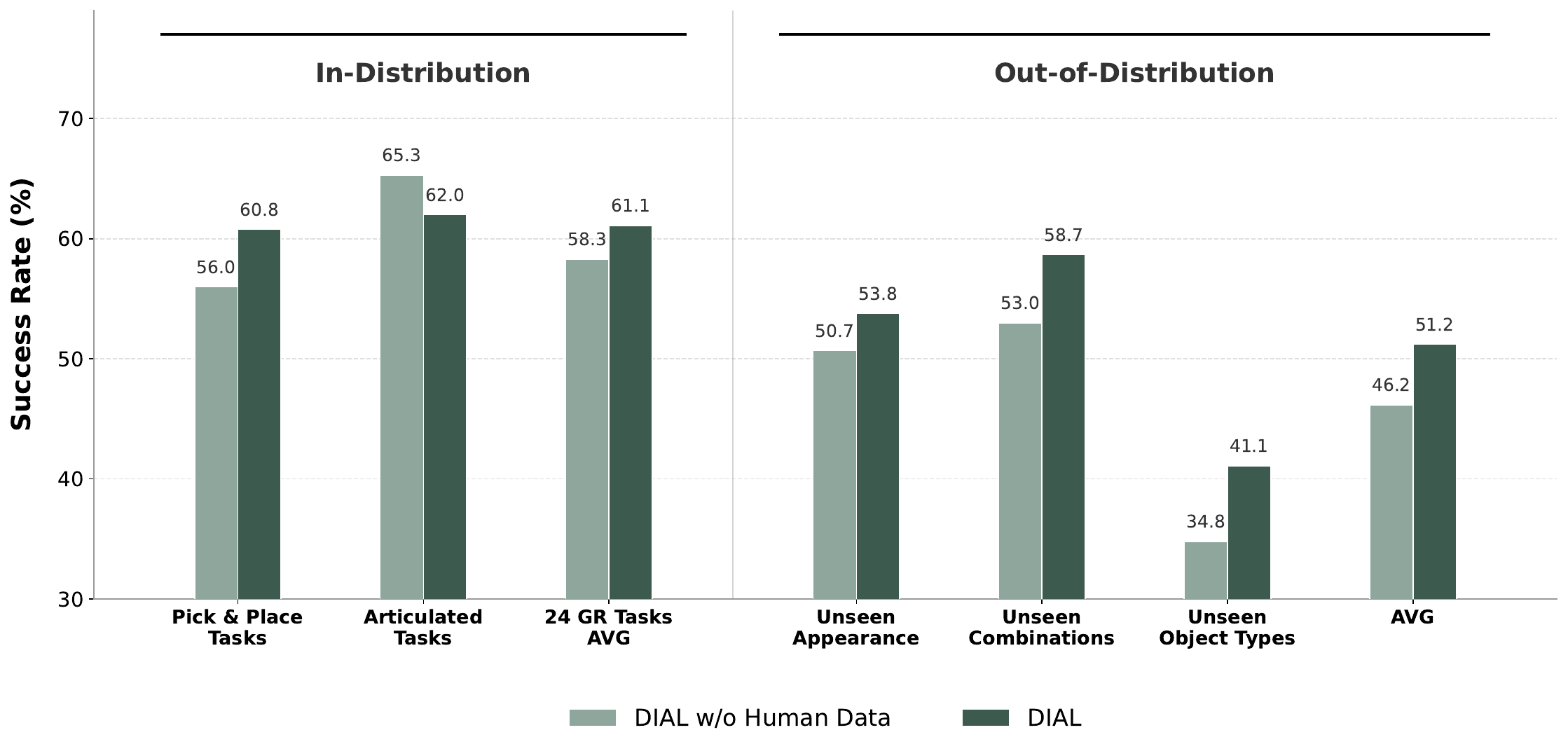}
    \caption{Impact of incorporating EgoDex \texttt{basic\_pick\_place} human demonstrations on few-shot performance in RoboCasa GR1 simulation tasks.}
    \label{fig:results_robocasa_egodex_narrow}
\end{figure}

We evaluate DIAL's ability to leverage cross-embodiment human demonstrations from the EgoDex \texttt{basic\_pick\_place} subset to improve both in-distribution performance and OOD generalization in RoboCasa GR1 simulation tasks. Results are detailed in Figure~\ref{fig:results_robocasa_egodex_narrow}.

\paragraph{In-Distribution Performance.} Incorporating human data yields a clear benefit for {Pick \& Place Tasks}, with success rates rising from {56.0\%} to {60.8\%}. This indicates that System-2 effectively distills the core logic of pick-and-place actions—such as reaching, grasping, and spatial placement—from diverse human demonstrations. In contrast, {Articulated Tasks} show no improvement (62.0\% with human data vs. 65.3\% without), primarily due to a {domain mismatch}: the EgoDex \texttt{basic\_pick\_place} subset contains only pure rearrangement interactions and lacks demonstrations involving articulated objects. As a result, the human-centric prior provides little guidance for this task category.

\paragraph{Out-of-Distribution Generalization.} The integration of human data has a pronounced impact on OOD performance, boosting zero-shot generalization across all three metrics. Success rates increase from {34.8\%} to {41.1\%} for \textit{unseen object types}, from {53.0\%} to {58.7\%} for \textit{unseen combinations}, and from {50.7\%} to {53.8\%} for \textit{unseen appearances}. These improvements suggest that exposure to diverse human-object interactions enables the VLM (System-2) to acquire a more robust semantic understanding of manipulatable objects, focus on abstract goals rather than specific object-container pairings, and handle novel visual appearances more effectively. Overall, the average OOD success rate rises from {46.2\%} to {51.2\%}, highlighting DIAL's strength in {cross-embodiment learning}.

Overall, these results indicate that DIAL scales effectively:  while the magnitude of gains depends on task coverage in the human dataset, integrating human demonstrations consistently enhances semantic reasoning and substantially improves OOD robustness.

% ----------------------------------------------------------------------
% Section 4.4: Addressing Q4
% ----------------------------------------------------------------------
\subsection{Real-World Robustness and Stability}
\label{sec:exp_real_world}

\begin{figure}[!t]
    \centering
    \includegraphics[width=\textwidth]{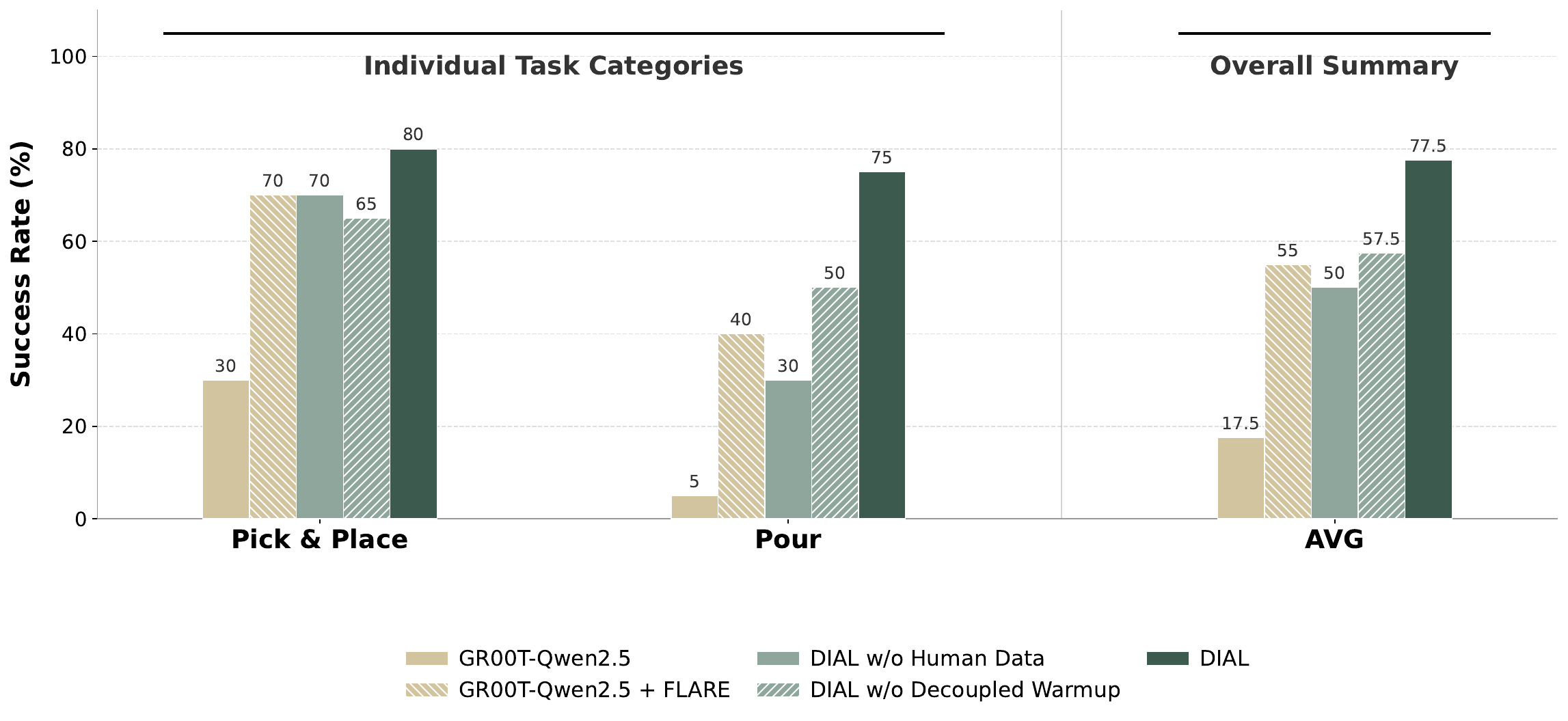}
    \caption{
    % In-distribution experiment results on the IRON-R01-1.11 robot. 
    In-distribution experiment results on the real-world humanoid robot for cross-embodiment learning tasks, showing the impact of joint training with human demonstrations. 
    % In-distribution results for cross-embodiment tasks. Performance on \textit{Pick \& Place} and \textit{Pouring} using the real-world humanoid robot, showing the impact of joint training with human demonstrations.
    }
    \label{fig:results_real_robot_id}
\end{figure}

\begin{figure}[!t]
    \centering
    \includegraphics[width=\textwidth]{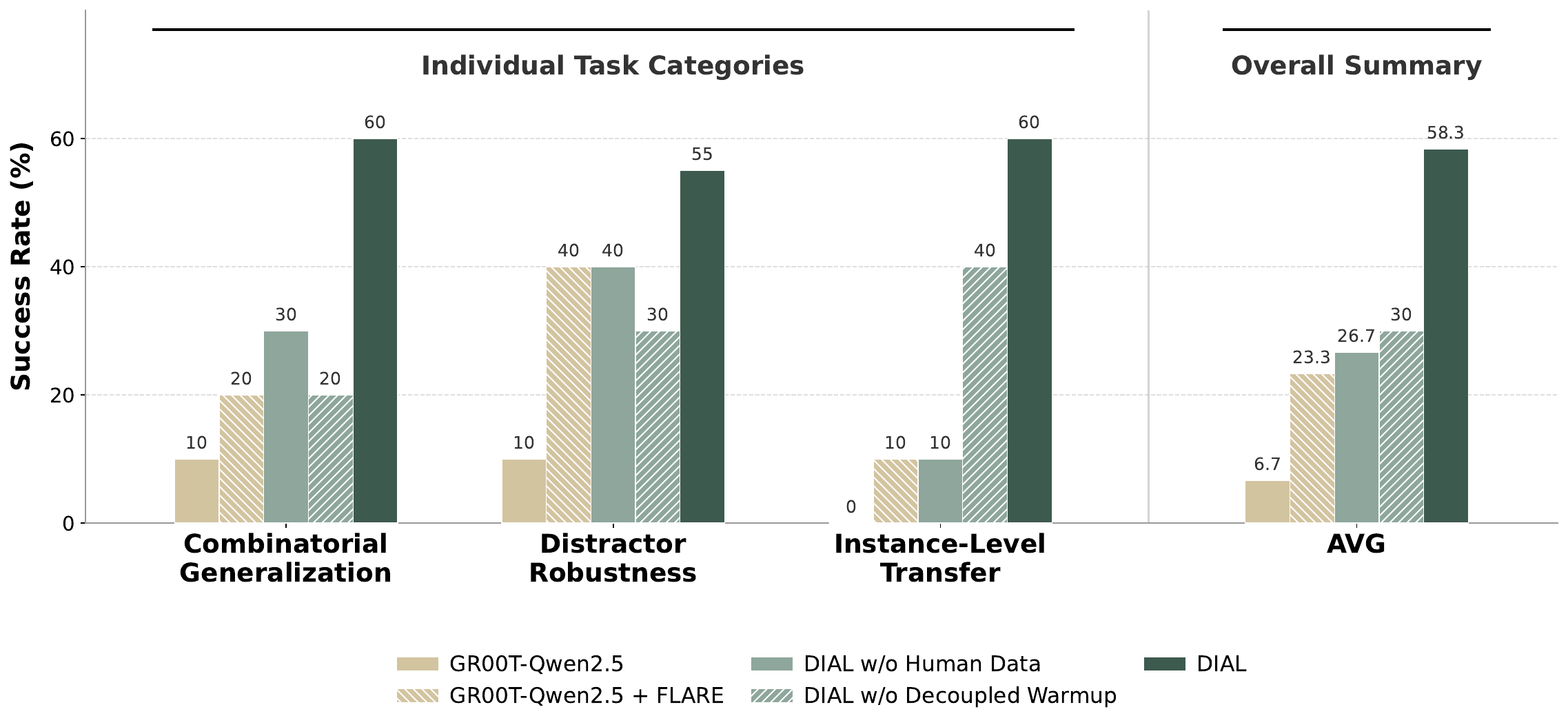}
    \caption{
    % Out-of-distribution experiment results on the IRON-R01-1.11 robot across three generalization 
    % challenges: combinatorial generalization, distractor robustness, and instance-level transfer. 
    Real-world OOD results for cross-embodiment learning tasks across three generalization challenges: combinatorial generalization, distractor robustness, and instance-level transfer. 
    }
    \label{fig:results_real_robot_ood}
\end{figure}

\begin{figure}[!t]
    \centering
    \includegraphics[width=\textwidth]{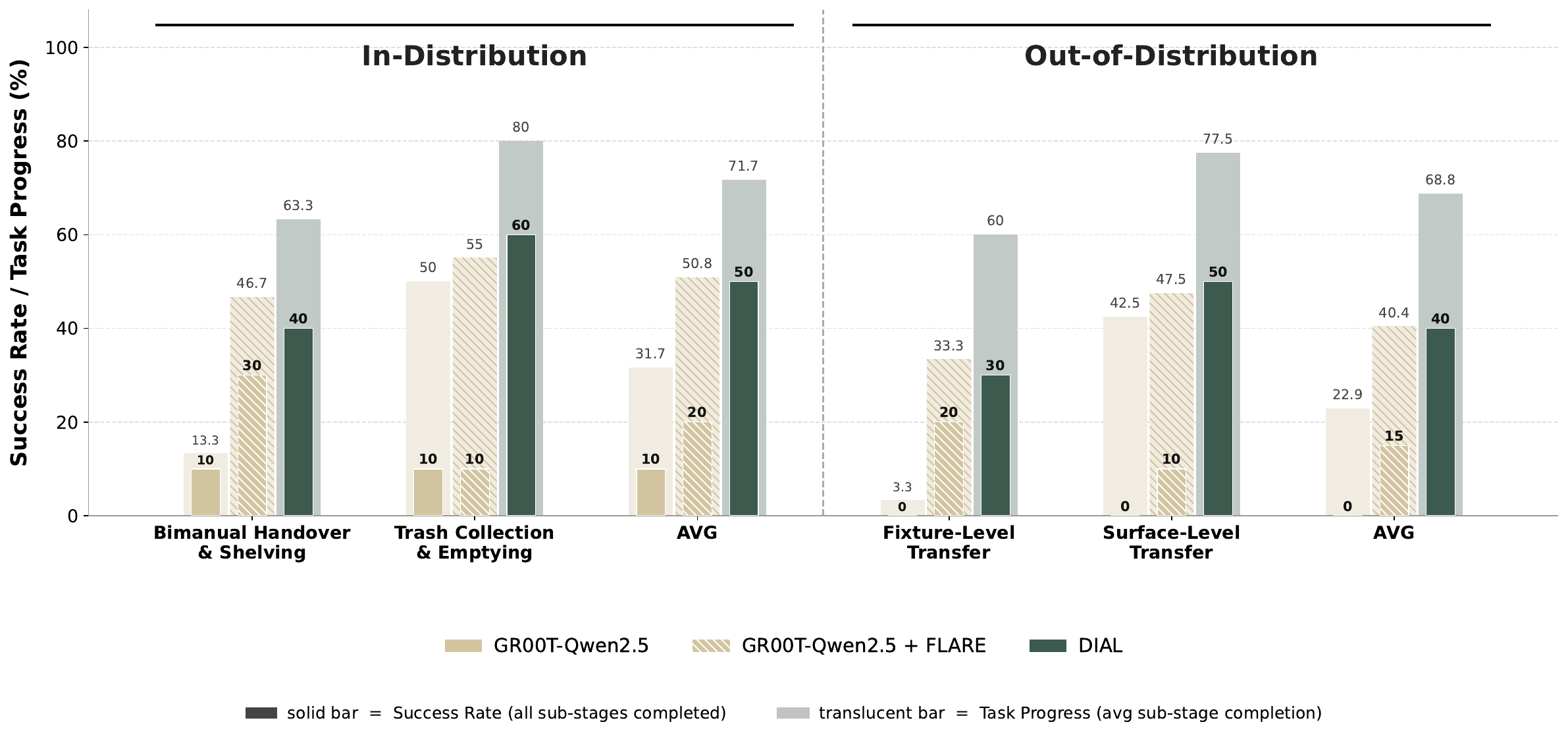}
    \caption{
    Real-world experiment results on multi-stage coordination tasks, evaluating both in-distribution performance and zero-shot generalization across novel fixture and surface configurations.
    }
    \label{fig:results_real_robot_multistage}
\end{figure}

% To further validate DIAL's physical viability and training stability, we deploy the model on the real-world IRON-R01-1.11 robot, with results shown in Figure~\ref{fig:results_real_robot_id} and Figure~\ref{fig:results_real_robot_ood}.

To further validate DIAL’s physical viability and training stability, we deploy the model on the real-world IRON-R01-1.11 robot across various task complexities, with results shown in Figure~\ref{fig:results_real_robot_id}--\ref{fig:results_real_robot_multistage}.

\paragraph{Importance of Decoupled Warmup.} During real-robot deployment, we observe that the decoupled warmup phase is crucial for the model's training stability. Ablation results indicate that removing this warmup leads to a severe drop in performance for both in-distribution tasks (average success rate drops from {77.5\%} to {57.5\%} in Figure~\ref{fig:results_real_robot_id}) and out-of-distribution scenarios (from {58.3\%} to {30.0\%} in Figure~\ref{fig:results_real_robot_ood}). Without warmup, System-2 fails to form coherent visual foresight before System-1 overfits to early noise, destabilizing joint optimization. The warmup phase allows System-1 to reliably track \textit{perfect} future states before confronting \textit{predicted} intentions, providing a stable foundation for generalization.

\paragraph{Generalization in Complex Environments.} Empowered by stable training, DIAL exhibits strong real-world generalization across several practical challenges (Figure~\ref{fig:results_real_robot_ood}). In \textit{combinatorial generalization} tests, despite the fine-tuning dataset containing only single-object pick-and-place trajectories, DIAL successfully identifies and manipulates the target object among multiple familiar objects based on language instructions, whereas baseline methods largely lose their instruction-following capabilities. In scenarios with \textit{unseen background distractors}, System-1 isolates the target by comparing the latent encoding of the current observation with System-2's predicted visual foresight, effectively filtering out background noise and preventing shortcut learning. In the more demanding \textit{instance-level transfer} setting, DIAL reliably grasps previously unseen bottles with different shapes and liquid colors to complete pouring tasks, demonstrating precise and adaptable control.

\paragraph{Role of Human Data.} We further find that incorporating human demonstrations during pre-training is essential for robust real-world performance. Removing human data significantly degrades OOD success rates (from {58.3\%} to {26.7\%} in Figure~\ref{fig:results_real_robot_ood}), highlighting the importance of diverse human priors for semantic reasoning and cross-embodiment generalization.

\textbf{Robustness in Multi-Stage Coordination.} Beyond basic manipulation tasks, DIAL demonstrates exceptional execution robustness in complex multi-stage coordination (Figure~\ref{fig:results_real_robot_multistage}). Without a structured foresight bottleneck, baselines lack a cognitive roadmap and struggle with phase transitions. For instance, they frequently get trapped in repetitive sweeping loops instead of transitioning to empty the dustpan, leading to severe performance degradation in both task progress and success rate. In contrast, DIAL’s latent intent explicitly envisions upcoming subgoals, providing a reliable temporal anchor that guides System-1 smoothly across intricate sub-stages. Furthermore, this robust grounding enables strong zero-shot generalization, allowing DIAL to successfully adapt to unseen multi-tier shelves (fixture-level transfer) and novel tablecloths (surface-level transfer).

% Overall, these real-robot results show that DIAL combines stable training with robust execution: decoupled warmup stabilizes joint optimization, human data enriches semantic understanding, and the system generalizes effectively across combinatorial, distractor, and instance-level challenges.

Overall, these real-robot results show that DIAL combines stable training with robust execution: decoupled warmup stabilizes joint optimization, human data enriches semantic understanding, predictive latent intent provides reliable guidance for long-horizon tasks, and the unified system generalizes effectively across diverse real-world configurations.

% and the system generalizes across diverse physical challenges, ranging from semantic disambiguation and object-level generalization to complex multi-stage coordination across novel workspace configurations.

% ----------------------------------------------------------------------
% Section 4.5: Addressing Q5
% ----------------------------------------------------------------------

\subsection{Interpreting Latent Foresight}
\label{sec:exp_viz}

\begin{figure}[!t]
    \centering
    \includegraphics[width=\textwidth, height=0.6\textheight]{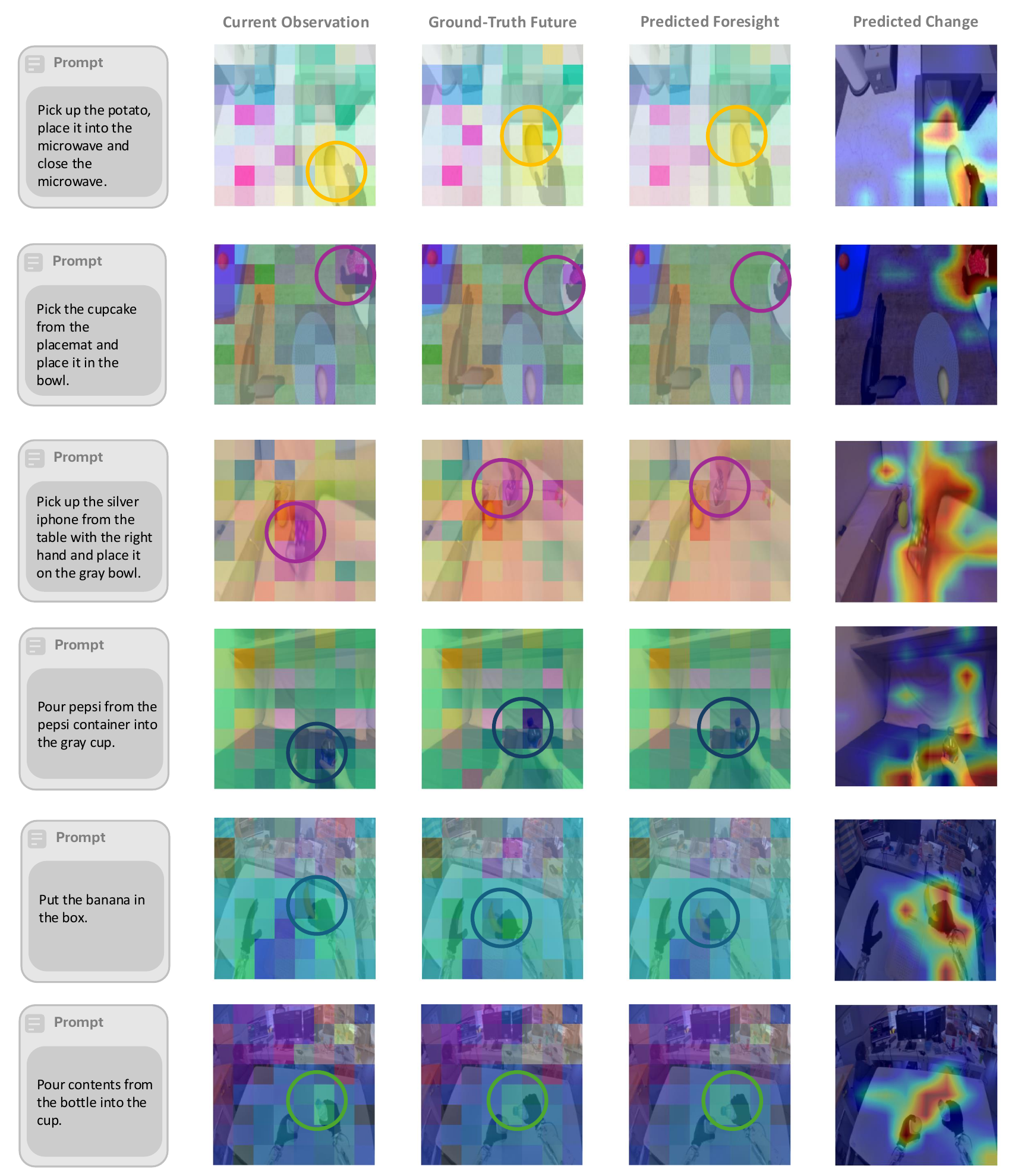}
    \caption{
    Visualization of latent representations for current observations, ground-truth futures, and predicted foresight, with colors encoding the first three PCA components mapped to RGB. 
    The last column shows the per-patch cosine distance between predicted foresight and current observation features, highlighting regions where the model anticipates future change.
    }
    \label{fig:latent_visualization}
\end{figure}

To understand the semantic structure of the information passed between System-2 and System-1, we qualitatively analyze the learned latent representations using Principal Component Analysis (PCA). By projecting the high-dimensional features of the current observation, the ground-truth future, and the predicted foresight onto their first three principal components (mapped to RGB channels), we can visually interpret the model's internal foresight.

As illustrated in Figure~\ref{fig:latent_visualization}, the spatial color distributions in the first three columns reflect the latent manifold structure. Taking the task \textit{``Pick up the silver iPhone''} (Row 3) as an example, the Predicted Foresight closely mirrors the Ground-Truth Future, especially in task-relevant regions such as the target object and destination container (highlighted by circles). Conversely, these predicted features diverge significantly from the Current Observation precisely in the areas where physical manipulation is expected to occur. The final column further illustrates these anticipated changes through the per-patch cosine distance between the predicted foresight and current observation features, with warmer colors indicating greater deviation.

This structural alignment between the predicted and ground-truth futures, coupled with their deliberate divergence from the initial state (evident in both the PCA color maps and the change heatmaps), proves that System-2 does not merely reconstruct the current scene. Instead, it actively anticipates meaningful state transitions within the semantic space. These visualizations confirm that DIAL's predictive bottleneck successfully serves as a spatially aligned and semantically grounded bridge: generating a coherent ``visual roadmap'' that System-1 subsequently decodes into precise physical actions.

\section{Conclusion and Discussion}

In this paper, we introduce \textbf{DIAL}, a novel framework for end-to-end VLA models that achieves a structural decoupling of cognitive decision making and motor execution through a differentiable {latent intent bottleneck}. By framing the VLM as a predictive latent world model (\textbf{System-2}) and the controller as a latent inverse dynamics model (\textbf{System-1}), DIAL ensures that every motor command is strictly grounded in the model's intent expressed by the latent visual foresight. Our extensive evaluations on the RoboCasa GR1 Tabletop benchmark and real-world deployments on the IRON-R01-1.11 humanoid robot demonstrate that DIAL establishes a new state-of-the-art in performance, achieving $10\times$ higher data efficiency than existing methods and exhibiting robust zero-shot generalization across novel objects and complex configurations, as well as maintaining stable execution in long-horizon, multi-stage coordination.

Looking forward, there are several key directions to further scale the DIAL framework. Currently, our System-1 employs a relatively small DiT backbone; scaling this to larger parameter sizes could significantly enhance the precision and multi-modal handling of complex motor tasks. Furthermore, while our current implementation keeps the {VLM-native ViT} frozen to maintain stable feature alignment, future work will explore {end-to-end fine-tuning} of this vision backbone, potentially stabilized by an {EMA-based encoding} strategy and {latent token compression} to further boost performance and efficiency.  Most importantly, since System-2 is designed for latent world modeling, DIAL is uniquely positioned to scale by consuming {massive, in-the-wild human videos} that lack action labels. Leveraging such action-free data to pre-train visual foresight will likely be the next frontier in building truly generalist embodied agents.

Ultimately, we envision a shift toward a more integrated yet modular paradigm for embodied intelligence. A compelling frontier is to incorporate {latent world modeling directly into the native pre-training tasks of foundational VLMs},  instilling actionable physical priors and a dynamics-oriented understanding within the backbone from the outset. This would ensure the VLM's representations are inherently aligned with the physical requirements of downstream control.  Furthermore, the decoupling of DIAL suggests a highly efficient iteration strategy: once a {System-1 action expert} is pre-trained to master motor control, new or updated VLM generations can be seamlessly coupled and aligned to it, enabling the rapid transfer of the latest cognitive advances to robotic embodiments. By treating latent foresight as the universal interface between reasoning and execution, DIAL paves the way for a new generation of versatile and scalable generalist agents.

% \clearpage

\medskip

\bibliography{neurips_2025}
\bibliographystyle{unsrt}

\end{document}